\documentclass[runningheads]{llncs}
\usepackage[T1]{fontenc}
\usepackage{hyperref}
\usepackage{url}
\usepackage{float}
\usepackage{comment}
\usepackage{subcaption}
\usepackage{multirow}
\usepackage{amsmath}
\usepackage{color}
\usepackage{latexsym,amssymb,graphicx,ntheorem}
\usepackage{amsfonts}

\usepackage[misc]{ifsym}

\begin{document}
\tocauthor 
\toctitle

\title{Learning Optimal Transport Between two Empirical Distributions with Normalizing Flows\thanks{This work was supported by the Artificial Natural Intelligence Toulouse Institute (ANITI, ANR-19-PI3A-0004), the AI Sherlock Chair (ANR-20-CHIA-0031-01), the ULNE national future investment programme (ANR-16-IDEX-0004) and the Hauts-de-France Region.}}
%Approximating optimal transport between two probability measures with normalizing flows \thanks{}}
%
\titlerunning{Learning optimal transport between two empirical distributions with NF}
\author{Florentin Coeurdoux\inst{1}\orcidID{0000-0002-2100-1438}\Letter \and Nicolas Dobigeon\inst{1}\orcidID{0000-0001-8127-350X} \and
Pierre Chainais\inst{2}\orcidID{0000-0003-4377-7584}}
\authorrunning{Coeurdoux et al.}
\institute{University of Toulouse, IRIT/INP-ENSEEIHT, F-31071 Toulouse, France \\
\email{\{Florentin.Coeurdoux, Nicolas.Dobigeon\}@irit.fr}\\ 
\and
Univ. Lille, CNRS, Centrale Lille, UMR 9189 CRIStAL, F-59000 Lille, France\\
\email{pierre.chainais@centralelille.fr}}
\maketitle              % typeset the header of the contribution
%
% - Parler de transport optimal 
% - Formulation approchée
% - on se restreint à la classe des NF 
% - apprentissage/optimisation
% - Perspective : Domain translation  
% - Prendre le temps d'être plus préci

\begin{abstract}
Optimal transport (OT) provides effective tools for comparing and mapping probability measures. We propose to leverage the flexibility of neural networks to learn an approximate optimal transport map. More precisely, we present a new and original method to address the problem of transporting a finite set of samples associated with a first underlying unknown distribution towards another finite set of samples drawn from another unknown distribution. We show that a particular instance of invertible neural networks, namely the normalizing flows, can be used to approximate the solution of this OT problem between a pair of empirical distributions. To this aim, we propose to relax the Monge formulation of OT by replacing the equality constraint on the push-forward measure by the minimization of the corresponding Wasserstein distance. The push-forward operator to be retrieved is then restricted to be a normalizing flow which is trained by optimizing the resulting cost function. This approach allows the transport map to be discretized as a composition of functions. Each of these functions is associated to one sub-flow of the network, whose output provides intermediate steps of the transport between the original and target measures. This discretization yields also a set of intermediate barycenters between the two measures of interest. Experiments conducted on toy examples as well as a challenging task of unsupervised translation demonstrate the interest of the proposed method. Finally, some experiments show that the proposed approach leads to a good approximation of the true OT.

\end{abstract}
\section{Introduction}

The optimal transport (OT) problem was initially formulated by the French mathematician Gaspard Monge. In his seminal paper published in 1781 \cite{monge}, he raised the following question: how to move a pile of sand to a target location with the least possible effort or cost? The objective was to find the best way to minimize this cost by a transport plan, without having to list all the possible matches between the starting and ending points. More recently, thanks to recent advances related to computational issues  \cite{cot}, OT has founded notable successes with respect to applications ranging from image processing and computer vision \cite{Sliced-OT} to machine learning \cite{CycleGAN-OT} and domain adaptation \cite{domain-adaptation-OT}.

Normalizing flows (NFs) have also attracted a lot of interest in the machine learning community, motivated in particular by their ability to model high dimensional data \cite{papamakarios2021normalizing,Kobyzev2020normalizing}. These deep generative models are characterized by an invertible operator that associates any input data distribution with a target distribution that is usually chosen to be Gaussian. They have the great advantage of leading to tractable distributions, which eases direct sampling and density estimation. Applications of these generative models include image generation with real-valued non-volume preserving transformations (RealNVP) \cite{realnvp} or generative flows using an invertible 1x1 convolution (GLOW) \cite{glow}.

Motivated by the similarities between the problem of OT and the training of NF, this paper proposes a neural architecture and a corresponding training strategy that permits to learn an approximate Monge map between any two empirical distributions. The proposed framework is based on a relaxation of the Monge formulation of OT. To adapt the training loss to the flow-based structure of the network, this loss function is supplemented with a Sobolev regularisation to promote minimal efforts achieved by each flow. Numerical simulations show that this regularisation results in a smoother and more efficient trajectory. Interestingly, the  discretization inherent to the flow-based structure of the network implicitly provides intermediate transports and, at the same time, Wasserstein barycenters \cite{agueh2011barycenters}. To the best of our knowledge, this is the first time that NFs are considered to address OT and Wasserstein barycenter computation, up to interesting dimensions. 

\textit{Contributions}. Our contributions are twofold: i) Section \ref{sec:TOC} recalls the Monge formulation of OT and proposes a relaxation in the case of a transport between two empirical distributions. ii) Section \ref{sec:NF} presents the generic framework based on NFs and describes a particular instance to solve the OT problem. Section \ref{sec:experiences} presents some experimental results illustrating the performance of the proposed method. Section \ref{sec:conclusion} concludes this paper.

\section{Relaxation of the optimal transport problem}\label{sec:TOC}

Let $\mu$ and $\nu$ be two probability measures with finite second order moments. More general measures, for example on $\mathcal{X}=\mathbb{R}^{d}$ (where $d \in \mathbb{N}^{*}$ is the dimension), can have a density $\mathrm{d} \mu(x)=p_{X}(x)dx$ with respect to the Lebesgue measure, often noted $p_{X}=\frac{\mathrm{d} \mu}{\mathrm{d} x}$, which means that
\begin{equation}
    \forall h \in \mathcal{C}\left(\mathbb{R}^{d}\right), \quad \int_{\mathbb{R}^{d}} h(x) \mathrm{d} \mu(x)= \int_{\mathbb{R}^{d}} h(x) p_{X}(x) \mathrm{d} x 
\end{equation}
where $\mathcal{C}(\cdot)$ is the class of continuous functions.
In the remainder of this paper, $\mathrm{d} \mu(x)$ and $p_{X}(x)\mathrm{d}x$ will be used interchangeably.

\subsection{Background on optimal transport}

Let consider $\mathcal{X}$ and $\mathcal{Y}$ two separable metric spaces. Any measurable application $T: \mathcal{X} \rightarrow \mathcal{Y}$ can be extended to the so-called push-forward operator $T_{\sharp}$ which moves a probability measure on $\mathcal{X}$ to a new probability measure on $\mathcal{Y}$. For any measure $\mu$ on $\mathcal{X}$, one defines the image measure $\nu=$ $T_{\sharp} \mu$ on $\mathcal{Y}$ such that
\begin{equation}
\forall h \in \mathcal{C}(\mathcal{Y}), \quad \int_{\mathcal{Y}} h(y) \mathrm{d} \nu(y)=\int_{\mathcal{X}} h(T(x)) \mathrm{d} \mu(x).
\end{equation}
Intuitively, the application $T: \mathcal{X} \rightarrow \mathcal{Y}$ can be interpreted as a function moving a single point from one measurable space to another \cite{cot}. The operator $T_{\sharp}$ pushes each elementary mass of a measure $\mu$ on $\mathcal{X}$ by applying the function $T$ to obtain an elementary mass in $\mathcal{Y}$. The problem of OT as formulated by Monge is now stated in a general framework. For a given cost function $c: \mathcal{X} \times \mathcal{Y} \rightarrow[0,+\infty]$, the measurable application $T: \mathcal{X} \rightarrow \mathcal{Y}$ is called the OT map from a measure $\mu$ to the image measure $\nu = T_{\#} \mu$ if it reaches the infimum
\begin{equation}
\inf _{T}\left\{\int_{\mathcal{X}} c(x, T(x)) \mathrm{d} \mu(x) : T_{\sharp} \mu=\nu \right\}.
\label{monge}
\end{equation}
Alternatively the Kantorovitch formulation of OT results from a convex relaxation of the Monge problem  \eqref{monge}. By defining $\Pi$ as the set of all probabilistic couplings %$\in \mathcal{P}\left(\mu \times \nu\right)$ 
with marginals $\mu$ and $\nu$, it yields the optimal $\pi$ that reaches
\begin{equation}\label{eq:Kantorovitch}
    \min_{\pi \in \Pi} \int_{\mathcal{X} \times \mathcal{Y}} c\left(\mathbf{x}, \mathbf{y}\right) d \pi\left(\mathbf{x}, \mathbf{y}\right)
\end{equation}
Under this formulation, the optimal $\pi$, which is a joint probability measure with marginals $\mu$ and $\nu$, can be interpreted as the optimal transportation map. It allows  the Wasserstein distance of order $p$ between $\mu$ and $\nu$ to be defined as
\begin{equation}
W_{p}\left(\mu, \nu\right) \stackrel{\text { def }}{=}\inf _{\pi \in \Pi}\left\{\left(\underset{\substack{\mathbf{x} \sim \mu \\ \mathbf{y} \sim \nu }}{\mathbb{E}} d\left(\mathbf{x}, \mathbf{y}\right)^{p}\right)^{\frac{1}{p}}\right\}
\end{equation}
where $d(\cdot,\cdot)$ is a distance defining the cost function $c\left(\mathbf{x}, \mathbf{y}\right)=d\left(\mathbf{x}, \mathbf{y}\right)^{p}$. The Wasserstein distance is also known as the Earth mover's distance. It defines a metric over the space of square integrable probability measures.

\subsection{Proposed relaxation of OT}

OT boils down to a variational problem, i.e., it requires the minimization of an integral criterion in a class of admissible functions. Given two probability measures $\mu$ and $\nu$, the existence and uniqueness of an operator $T$ that belongs to the class of bijective, continuous and differentiable functions such that $T_\sharp\mu = \nu$ is not guaranteed. The difficulty lies in the class defining these admissible functions. Indeed, even when $\mu$ and $\nu$ are regular densities on regular subsets of $\mathbb{R}^d$, the search for a transport map such that $T_{\sharp} \mu=\nu$ makes the problem \eqref{monge} difficult in a general case. To overcome the difficulty of solving this equation on $T_\sharp$, we propose to reformulate the Monge's OT statement by relaxing the equality on the operator defining the image measure. %\\

More precisely, the equality between the image measure $T_{\sharp} \mu$ and the target measure $\nu$ is replaced by the minimization of their statistical distance $d(T_{\sharp}\mu,\nu)$. The choice of the distance $d(\cdot,\cdot)$ is crucial because it determines the quality of the approximation of the image measure by the transport map $T$. In this work, we propose to choose $d(\cdot,\cdot)$ as the Wasserstein distance ${W}_{p}(\cdot,\cdot)$. This choice will be motivated by the fact that this distance can be easily approximated numerically without explicit knowledge of the probability distributions $\mu$ and $\nu$, in particular when they are empirically described by samples only. The relaxation of the Monge problem \eqref{monge} can then be written as
\begin{equation}
\inf _{T}\left\{{W}_{p}(T_{\sharp}\mu, \nu) + \lambda \int_{\mathcal{X}} c(x, T(x)) \mathrm{d} \mu(x)\right\}
\label{relaxed}
\end{equation}
where the cost function defined in \eqref{monge} is interpreted here as a regularisation term adjusted by the hyperparameter $\lambda$.

\begin{remark}
The relaxed formulation \eqref{relaxed} relies on the Wasserstein distance between the target measure $\nu$ and the image measure $T_{\sharp}\mu$. This term should not be confused with the Wasserstein distance ${W}_{p}(\mu, \nu) $ which is the infimum reached by the solution of the Kantorovitch's formulation of OT \eqref{eq:Kantorovitch}.
\end{remark}

\subsection{Discrete formulation}\label{subsec:discrete_problem}
In a machine learning context, the underlying continuous measures are conventionally approximated by empirical point measures thanks to available data samples. Therefore, in this paper, we are interested in discrete measures and the empirical formulation of the OT problem. Within this framework, we will consider $\mu$ and $\nu$ two discrete measures described by the respective samples $\mathbf{x}=\left\{x_n\right\}_{n=1}^N$ and $\mathbf{y}=\left\{y_n\right\}_{n=1}^N$ such that ${\mu}=\frac{1}{N}\sum_{n=1}^{N}  \delta_{x_{n}}$ and ${\nu}=\frac{1}{N}\sum_{n=1}^{N}  \delta_{y_{n}}$. In the following, an empirical version of the criterion \eqref{relaxed} is proposed in the case of discrete measures. 

The formulation \eqref{relaxed} requires the evaluation of a Wasserstein distance whose computation is not trivial in its original form, especially in high dimension. An alternative consists in considering its rewriting in the form of the \emph{sliced-Wasserstein} (SW) distance. The idea underlying the SW distance is to represent a distribution defined in high dimension thanks to a set of projected one-dimensional distributions for which the computation of the Wasserstein distance is closed-form. Let $p_X$ and $p_Y$ denote the probability distributions of the random variables $X$ and $Y$. For any vector on the unit sphere $u \in \mathbb{S}^{d-1}$, the projection operator $S_u :\mathbb{R}^d \rightarrow \mathbb{R}$ is defined as $S_{u}(x) \triangleq\langle u, x\rangle$. The SW distance of order $p \in [1, \infty)$ between $p_X$ and $p_Y$ can be written \cite{swd}
\begin{equation}
SW_{p}\left(p_{X}, p_{Y}\right)=\left(\int_{\mathbb{S}^{d-1}} {W}_{p}\left(S_{u \sharp} p_{X}, S_{u \sharp} p_{Y}\right)^{p} d u \right)^{\frac{1}{p}}
\end{equation}
where the distance $W_p(\cdot,\cdot)$ defining the integrand is now one-dimensional, leading to an explicit computation by inversion of the cumulative distribution functions. In the case where the distributions $p_X$ and $p_Y$ are represented by the respective samples $\mathbf{x}$ and $\mathbf{y}$, a numerical Monte Carlo approximation of the SW distance  is
\begin{equation}\label{eq:SW_MonteCarlo}
\widehat{SW}_{p}(\mathbf{x}, \mathbf{y}) = \frac{1}{J} \sum_{j=1}^{J} {W}_{p}\left(\frac{1}{N} \sum_{n=1}^{N} \delta_{S_{u_j}({x}_{n})}, \frac{1}{N} \sum_{n=1}^{N} \delta_{S_{u_j}({y}_{n})}\right)
\end{equation}
where $u_1,\ldots,u_J$ are drawn uniformly on the sphere $\mathbb{S}^{d-1}$. The empirical form of the relaxation of the Monge problem \eqref{relaxed} is then written as
\begin{equation}
\min_{T}\left\{\widehat{SW}_{p}(T(\mathbf{x}), \mathbf{y}) + \lambda \sum_{n=1}^N c\left(x_{n}, T\left(x_{n}\right)\right) \right\}
\label{relaxe}
\end{equation}
where, with a slight abuse of notations, $T(\mathbf{x}) \triangleq \{T(x_n)\}_{n=1}^N$. 

\section{Normalizing flows to approximate OT}\label{sec:NF}

This section proposes to solve the problem \eqref{relaxe} by restricting the class of the operator $T$ to a class of invertible deep networks referred to as normalisation flows. The structure and the main properties of these networks are detailed in paragraph \ref{subsec:NF}. The strategy proposed to train these networks to solve the problem \eqref{relaxe} is then detailed in paragraph \ref{subsec:training}.

\subsection{Normalizing flows}\label{subsec:NF}
Normalization flows are a flexible class of deep generative networks that intend to learn a change of variable between two probability distributions $p_{X}$ and $p_{Y}$ through an invertible transformation $T_{\boldsymbol{\Theta}}: X \mapsto Y=T_{\boldsymbol{\Theta}}(X)$ parametrized by $\boldsymbol{\Theta}$. In general, the distribution $p_X$ is only known through samples $\mathbf{x}=\left\{x_n\right\}_{n=1}^N$ and, for tractability purpose, the distribution $p_Y$ is chosen as a centered normal distribution with unit variance. The parameters $\boldsymbol{\Theta}$ defining the operator $T_{\boldsymbol{\Theta}}$ are then adjusted by maximizing the likelihood associated with the observations $\mathbf{x}$ according to the change of variable formula 
\begin{equation}
p_{X}(x)=p_{Y}\left(T_{\boldsymbol{\theta}}(x)\right)\left|\operatorname{det} J_{T_{\boldsymbol{\theta}}^{-1}}\right|
\label{change_var}
\end{equation}
with $J_{T_{\boldsymbol{\Theta}}^{-1}}=\frac{\partial T_{\boldsymbol{\theta}}^{-1}}{\partial x}$. NF networks obey a cell-like structure, explicitly defining the operator $T_{\boldsymbol{\Theta}}(\cdot)$ as the composition of $M$ functions $T^{(m)}_{\boldsymbol{\theta}_m}$, usually referred to as \emph{flows}, i.e.,
\begin{equation}\label{eq:layered_structure}
    T_{\boldsymbol{\Theta}}(\cdot)=T^{(M)}_{\boldsymbol{\theta}_M} \circ T^{(M-1)}_{\boldsymbol{\theta}_{M-1}}\circ \ldots \circ T^{(1)}_{\boldsymbol{\theta}_1}(\cdot)
\end{equation}
with ${\boldsymbol{\Theta}}= \left\{\boldsymbol{\theta}_1,\ldots,\boldsymbol{\theta}_M\right\}$. In the following, to lighten notations, each sub-function contributing to the flow will be denoted by $T_m=T^{(m)}_{\boldsymbol{\theta}_m}$. In the present work, these functions are chosen as coupling layers as implemented by flows like RealNVP \cite{realnvp} and nonlinear independent component estimation (NICE) \cite{NICE}. These coupling layers ensure an invertible transformation and an explicit expression of the Jacobian required in the change of variables \eqref{change_var}. The input and output of the $m$th layer are related as $\left(y_{\text {id }}, y_{\text {ch}}\right) = T_m\left(x_{\text {id }}, x_{\text {ch}}\right)$ with
%\begin{equation}
%\left(y_{\text {id }}, y_{\text {ch}}\right) = T_m\left(x_{\text {id }}, x_{\text {ch}}\right)
%\end{equation}
%with
\begin{equation}
\begin{cases}
y_{\text {id }}=x_{\text {id }} \\ 
y_{\text {ch}}=\left(x_{\text {ch}}+\mathsf{D}_m\left(x_{\text {id }}\right)\right) \odot \exp \left(\mathsf{E}_m\left(x_{\text {id }}\right)\right)
\end{cases}    
\label{coupling}
\end{equation}
where $x_{\text{id }}$ and $x_{\text{ch}}$ (resp. $y_{\text{id }}$ and $y_{\text{ch}}$) are disjoint subsets of components of the input vector $x$ (resp. the output vector $y$). The splitting of the input $x$ into $x_{\mathrm{id}}$ and $x_{\text{ch}}$ is achieved by a masking process such that $x_{\text{ch}}=\textrm{mask}(x)$ is transformed into a function of the unchanged part $x_{\text {id }}$. The scale function $\mathsf{E}_m(\cdot)$ and the offset function $\mathsf{D}_m(\cdot)$ are then described by neural networks whose parameters $\boldsymbol{\theta}_{m}$ need to be adjusted during the training. It is worth noting that imposing the flow-based architecture detailed in \eqref{eq:layered_structure} will lead to an explicit discretization scheme of the transport map $T_{\boldsymbol{\Theta}}(\cdot)$ into a sequence of elementary transport functions $T_m(\cdot)$. As it will be shown in Section \ref{subsec:barycenter}, this discretization has the great advantage of providing Wasserstein barycenters associated with the two measures $\mu$ and $\nu$. Note that the proposed method is not limited to NFs composed of coupling layers such as RealNVP \cite{realnvp}, NICE \cite{NICE} or GLOW \cite{glow}. It can be generalized to other types of NFs, including free-form Jacobian of reversible dynamics (FFJORD) \cite{FFJORD} and masked autoregressive flows (MAF) \cite{maf}.

\subsection{Loss function}\label{subsec:training}

As mentioned before, the objective of this work is to learn a bijective operator relating any two distributions $p_X$ and $p_Y$ described by samples $\mathbf{x}$ and $\mathbf{y}$. The search for this operator is restricted to the class of invertible deep networks $T_{\boldsymbol{\Theta}}$ described in paragraph \ref{subsec:NF}. The conventional strategy to train the network would be to maximize the likelihood defined by \eqref{change_var}. However this approach cannot be implemented in the context of interest here since the base distribution $p_Y$ is no longer explicitly given: it is only available through the knowledge of the set of samples $\mathbf{y}$. As a consequence, to adjust the weights of the network, the proposed alternative interprets the underlying learning task as the search for a transport map. Then a first idea would be to adjust these weights by directly solving the problem \eqref{relaxe}. However, to take advantage of the flow-based architecture of the  operator $T_{\boldsymbol{\Theta}}(\cdot)$, it seems legitimate to equally distribute the transport efforts provided by each flow. Thus, the regularization in \eqref{relaxe} will be instantiated for each elementary transformation $T_m(\cdot)$ associated to each flow of the network. %\\ 

\begin{figure}
    \centering
    \includegraphics[width=0.7\linewidth]{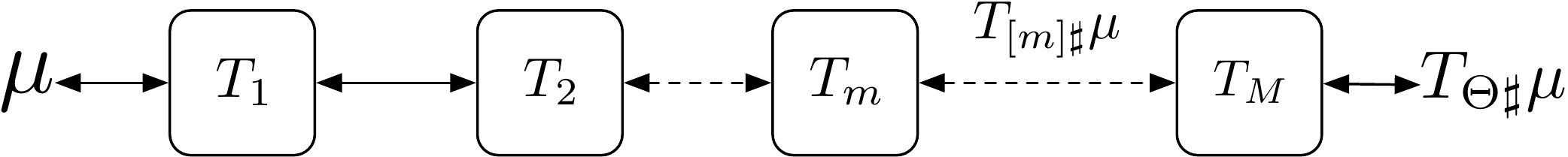}
    \caption{Architecture of the proposed SWOT-Flow.}
    \label{fig:SWOT-Flow}
\end{figure}

Moreover, when fitting deep learning-based models a major challenge arises from the stochastic nature of the optimization procedure, which imposes to use partial information (e.g., as mini-batches) to infer the whole structure of the optimization landscape. On top of that, the cost function to be optimized is not numerically constant since the approximation $\widehat{SW}$ of the SW distance in \eqref{relaxe} depends on the precise set of random vectors $\left\{{u}_{j}\right\}_{j=1}^{J}$ drawn over the unit sphere. To alleviate these optimization difficulties, we propose to further regularize the objective function by penalizing the energy $\left|{J_{T_m}(\cdot)}\right|^2$ of the Jacobians associated with the transformations $T_m(\cdot)$, $m=1,\ldots,M$. These Sobolev-like penalties promote regular operators $T_m(\cdot)$, promoting an overall operator $T_{\boldsymbol{\theta}}(\cdot)$ regular itself \cite{hoffman2020robust}. In the context of optimal transport, this regularization has already been studied in depth in \cite{jean_louet}. In that work, the author focused on the penalization of the Monge's formulation of OT by the $\ell_2$-norm of the Jacobian. It stated the existence of an optimal transport map $T$ solving the minimization problem
\begin{equation}
\inf_T \left\{\int_{\mathcal{X}}\left(|T(x)-x|^{2}+\gamma |J_{T}|^{2}\right) T(x) \mathrm{d} x: T_{\#} \mu=\nu\right\}   
\end{equation}
This formulation of OT imposes the transport map $T$ to be regular rather than deducing its regularity from its optimal properties. Finally, the training of the NF is carried out by minimizing the loss function
\begin{equation}
   \underbrace{\widehat{SW}_{p}(\mathbf{x}, \mathbf{y})}_\text{SW} + \underbrace{\sum_{n=1}^{N} \sum_{m=1}^{M}  \bigg[ \lambda c(T_{m-1}(x_n), T_m(x_n)) + \gamma \left|{J_{T_m}}(x_n)\right|^{2} \bigg]}_\text{Reg} 
   \label{eq:loss_function}
\end{equation}
with $T_0(x_n) = x_n$. The proposed network, whose general architecture is depicted in Fig. \ref{fig:SWOT-Flow}, will be referred to as SWOT-Flow in what follows. 

\subsection{Intermediate transports and Wasserstein barycenters}\label{subsec:barycenter}
As a consequence of the multiple-flow architecture \eqref{eq:layered_structure} of the NF, the transport map operated by the proposed SWOT-Flow is a composition of the $M$ individual flows $T_{m}(\cdot)$ ($m=1,\ldots,M$). Thus each flow implements an elementary transport and the composition of the first $m$ flows defined as
\begin{equation}
    T_{[m]}(\cdot) \triangleq T_{m}\circ \ldots \circ T_{1} (\cdot)
\end{equation}
can be interpreted as an intermediate step of the transport map from the input measure $\mu$ towards the target measure $\nu$, with $T_{[M]}(\cdot) \triangleq T_{\boldsymbol{\Theta}}(\cdot)$. Interestingly, these intermediate transports can be related to Wasserstein barycenters between $\mu$ and $\nu$ defined by \cite{agueh2011barycenters}
\begin{equation}
\label{eq:barycenter}
    \inf_{\beta} \left\{\alpha W_p(\mu,\beta) + (1-\alpha) W_p(\beta,\nu)\right\}.
\end{equation}
Indeed, the next section dedicated to numerical experiments will empirically show that  $T_{[m]\sharp}\mu$  approaches the solution of the problem \eqref{eq:barycenter} for the specific choice of the weight $\alpha = \frac{m}{M}$. In other words, the image measures provided by each intermediate transport operated by SWOT-Flow, i.e., as the outputs of each of the $M$ flows, can legitimately be interpreted as Wasserstein barycenters. 

\section{Numerical experiments}\label{sec:experiences}

This section assesses the versatility and the accuracy of SWOT-Flow through two sets of numerical experiments. First, several toy experiments are presented to provide some insights about key ingredients of the proposed approach. Then the performance of SWOT-Flow is illustrated through the more realistic and challenging task of unsupervised alignment of word embeddings in natural language processing. The source code is publicly available on GitHub \footnote{\href{https://github.com/FlorentinCDX/SWOT-Flow/}{FlorentinCDX/SWOT-Flow}}.

\subsection{Toy examples}\label{sec:toy_examples}
In these experiments, the proposed framework SWOT-Flow is implemented and tested with synthetic data. In all experiments, the input distributions are described by the respective samples $\mathbf{x}=\left\{x_n\right\}_{n=1}^N$ and $\mathbf{y}=\left\{y_n\right\}_{n=1}^N$ such that ${\mu}=\frac{1}{N}\sum_{n=1}^{N}  \delta_{x_{n}}$ and ${\nu}=\frac{1}{N}\sum_{n=1}^{N}  \delta_{y_{n}}$ with $N=20000$. The cost function $c(\cdot,\cdot)$ is chosen as the squared Euclidean distance, i.e., $c(x, y)=\|x-y\|_{2}^{2}$. However, it is worth noting that the proposed method is not limited to this Euclidean distance and can handle other costs defined on $\mathbb{R}^{d}$ or even on curved domains. 

\subsubsection{Implementation details.}
\begin{figure}
    \centering
    \includegraphics[width=0.65\linewidth]{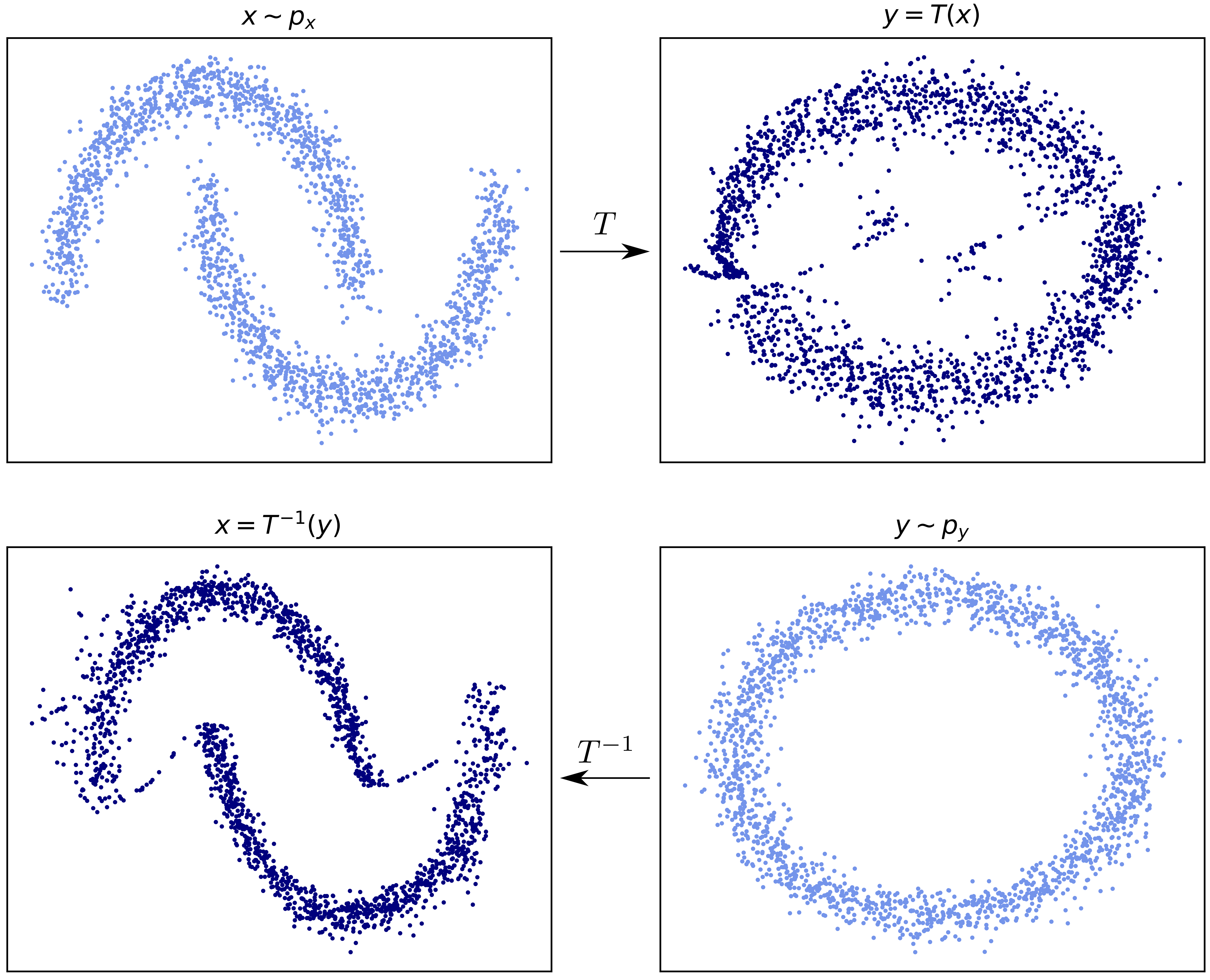}
    \caption{Operator $T$ learnt by SWOT-Flow when the base distribution $p_X$ is a double-moon (top left) and the target distribution $p_Y$ is a circle (bottom right).}
    \label{fig:push-foward}
\end{figure}

The stochastic gradient descent used to solve \eqref{eq:loss_function} is implemented in Pytorch. We use Adam optimizer with learning rate $10^{-4}$ and a batch size of $4096$ or $8192$ samples. The NF implementing $T_{\boldsymbol{\Theta}}(\cdot)$ is a RealNVP \cite{realnvp} for the example of Fig.~\ref{fig:push-foward} and an ActNorm type architecture network \cite{glow} for Fig.~\ref{fig:barycenter} and Fig.~\ref{fig:transport-gen}. It is composed of $M=4$ flows, each composed of two four-layer neural networks corresponding to $\mathsf{D}_m(\cdot)$ and $\mathsf{E}_m(\cdot)$ ($d \rightarrow 8 \rightarrow 8 \rightarrow d$) using hyperbolic tangent activation function. During training,  the number $J$ of slices drawn to approximate the SW distance  in \eqref{eq:SW_MonteCarlo} has been progressively increased, starting from $J=500$ to $J=2000$ by step of $50$ slices. At each epoch, new slices are uniformly drawn over the unit sphere and 100 epochs are carried out for each number of slices. The training procedure consist in 1) defining the loss function as the sole SW term in \eqref{eq:loss_function} from $J=500$ to $1500$ slices and then 2) incorporating the regularization term denoted as Reg in \eqref{eq:loss_function} where hyperparameters $\lambda$ and $\gamma$ are increased by a factor of $5\%$ every step of 100 slices.

%%% Figure double-moon -> circle
%%%

\subsubsection{Qualitative results.}
As a first illustration of the flexibility of the proposed approach, Fig.~\ref{fig:push-foward} shows the results obtained after learning an operator $T$ that transports a double moon-shaped distribution $p_X$ (top left) to a circle-shaped distribution (bottom right). The empirical image measures $T_\sharp p_X$ (top right) and $T^{-1}_\sharp p_Y$ (bottom left) are obtained by applying the estimated $T(\cdot)$ operator or its inverse $T^{-1}(\cdot)$. It is worth noting that the difficulty inherent to this experiment lies in the respective disjoint and non-disjoint supports of the two distributions. Despite the regularity of the trained NF, a very good approximation of the OT is learnt, even in presence of this topological change.

%% Fig. barycenters cicrle, rectangle
\newcommand{\subfigsize}{0.24\linewidth}
\begin{figure}
    \centering
    \includegraphics[width=\subfigsize, height=\subfigsize]{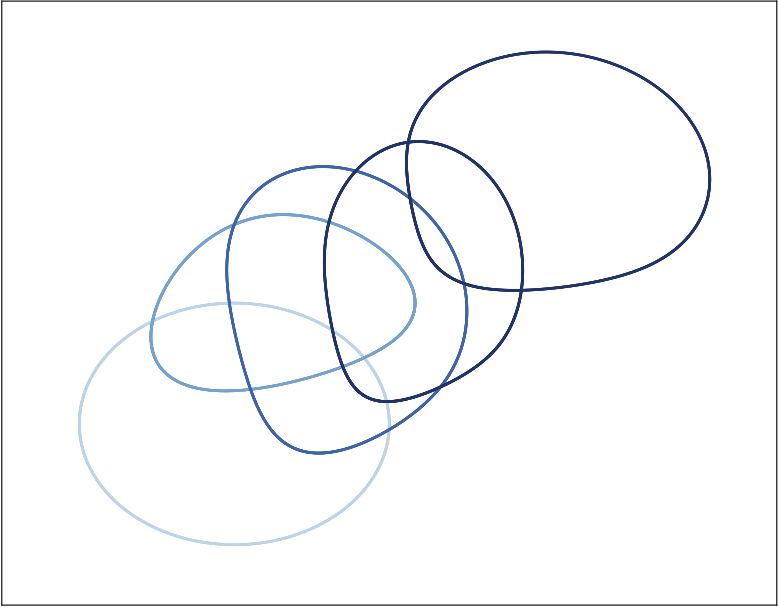}%
    \includegraphics[width=\subfigsize, height=\subfigsize]{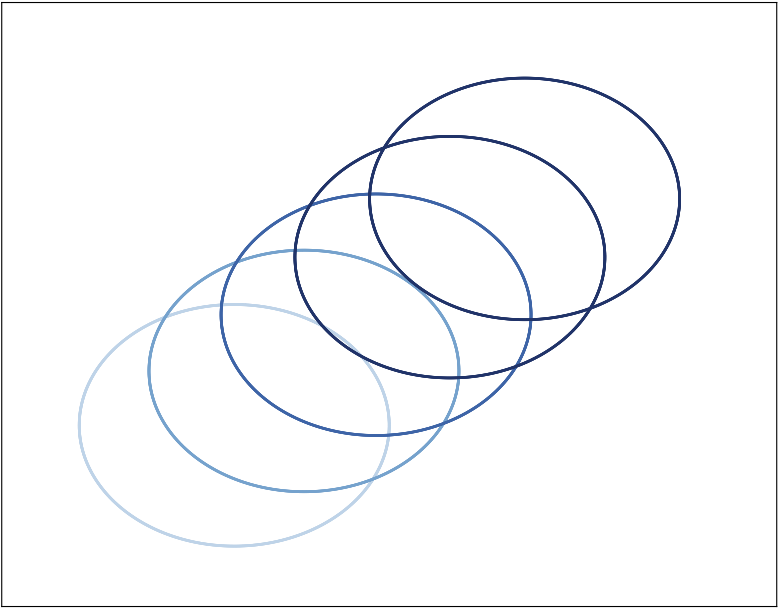}
    \includegraphics[width=\subfigsize, height=\subfigsize]{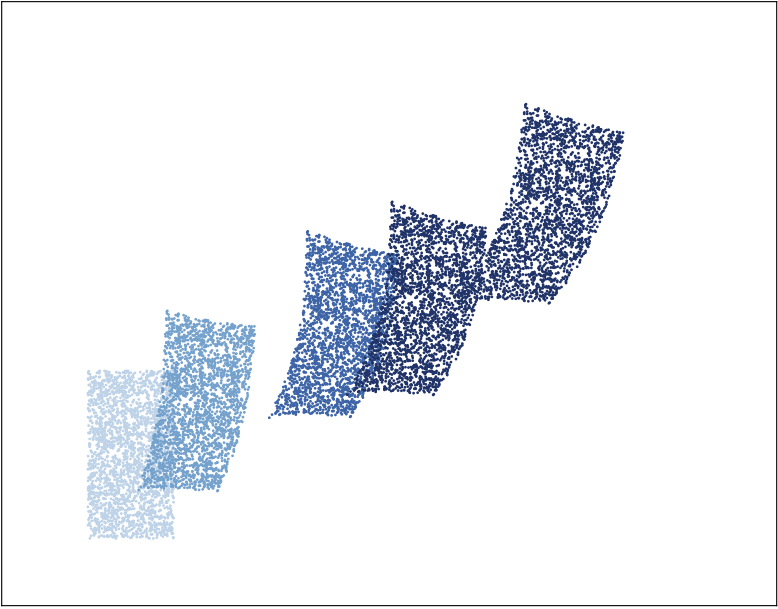}%
    \includegraphics[width=\subfigsize, height=\subfigsize]{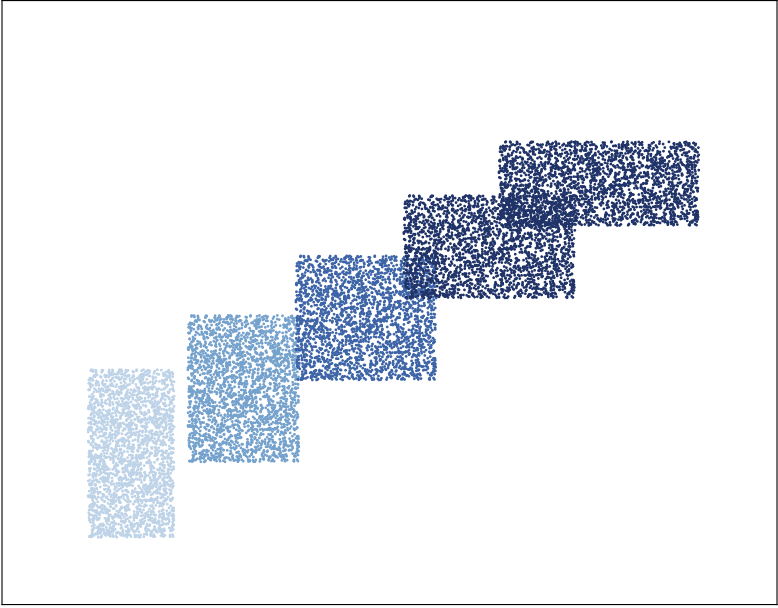}
    \caption{Elementary transports achieved by the proposed NF when trained without (1st and 3rd panels) or with (2nd and 4th panels) the regularization for protocols $\mathcal{P}_1$ (left panels) and $\mathcal{P}_2$ (right panels).}
    \label{fig:barycenter}
\end{figure}
%%%%

Fig. \ref{fig:barycenter} aims at illustrating the relevance of the Sobolev-like regularization (i.e., the $\ell_2$-norm of the Jacobian) included into the loss function \eqref{eq:loss_function} defined to train the NF. The first simulation protocol considers circle-shaped distributions while the second case considers rectangle-shaped distributions. In what follows, these two cases will be referred to as $\mathcal{P}_{1}$ and   $\mathcal{P}_{2}$, respectively. In this experiment, the objective is to learn the transport map from an initial distribution $p_X$ (light blue) to a target distribution $p_Y$  (dark blue) which is translated for $\mathcal{P}_1$ and both translated and stretched for $\mathcal{P}_2$. The color gradient shows the outputs of the $M$ successive flows of the network, i.e. the image measures $T_{[m]\sharp}p_X$ for $m=1,\ldots,M$. In the absence of regularization (left), the successive elementary transports clearly suffer from multiple unexpected deformations (superfluous translations and dilations). In contrast, when the loss is complemented with the proposed Sobolev-type penalty (right), the learnt operator $T$ is decomposed as a sequence of much more regular elementary transports. The resulting transport appears to be very close to optimal. In case $\mathcal{P}_{1}$, the expected translation is recovered, as well as the combined translation and stretching in case $\mathcal{P}_{2}$.

%% Table toy examples
\begingroup
\setlength{\tabcolsep}{7pt} % Default value: 6pt
\renewcommand{\arraystretch}{1.15} % Default value: 1
\begin{table}[h!]
\caption{Overall cost $\bar{C}$ and elementary costs  $\bar{c}_m$ required by each flow $T_m(\cdot)$ of the NF trained with or without (w/o) regularization for protocols $\mathcal{P}_1$ (circle-shaped  distributions) and $\mathcal{P}_2$ (rectangle-shaped distributions).}
\begin{center}
\begin{tabular}{|cc|ccccc|}
    \cline{3-7}
    \multicolumn{2}{c|}{} & \multicolumn{1}{c}{$\bar{c}_1$} & \multicolumn{1}{c}{$\bar{c}_2$} & \multicolumn{1}{c}{$\bar{c}_3$} & \multicolumn{1}{c}{$\bar{c}_4$} & \multicolumn{1}{c|}{$\bar{C}$}  \\
    \hline
    \multirow{2}{*}{$\mathcal{P}_1$}& w/o regularization  & 150.13  & 110.94  & 108.41   & 151.65 &  521.12   \\
    &with regularization  & 90.20       & 90.70       & 90.71       & 90.22  & 361.22\\
    \hline
    \multirow{2}{*}{$\mathcal{P}_2$}& w/o regularization  & 154.99       & 98.67       & 52.49   & 101.21 &  407.38   \\
    & with regularization & 88.77       & 89.42      & 89.43       & 89.38  & 357.0 \\
    \hline
\end{tabular}
\end{center}
\label{tab:cost}
\end{table}
\endgroup

To be more precise quantitatively, Table \ref{tab:cost} compares some metrics obtained when the NF has been trained using the regularization-free or regularized loss function, as defined in \eqref{eq:loss_function}. For the two aforementioned simulation protocols, it reports the elementary costs 
\begin{equation}
    \bar{c}_m = \frac{1}{N} \sum_{n=1}^N \left\|T_{m-1}(x_n)-T_{m}(x_n)\right\|_{2}^{2} 
    \label{eq:transport-cost}
\end{equation}
spent by each of the M flows $T_1(\cdot),\ldots,T_M(\cdot)$ to achieve the transport maps retrieved by SWOT-Flow. This table (last column) also reports the overall cost $\bar{C} =  \sum_{m=1}^M \bar{c}_m$. For the two simulation protocols $\mathcal{P}_{1}$ and $\mathcal{P}_{2}$, these results clearly show cheaper transports when using the proposed regularization. For instance, for the simulation protocol $\mathcal{P}_1$, the overall cost is $\bar{C} =360$ with the regularization, compared to $\bar{C} =520$ when it is omitted. Moreover, when using the regularized loss function, this cost is distributed homogeneously over the successive flows, with a variation of at most $\pm 1\%$ from one flow to another, against $ \pm 20\%$ otherwise.

%% Figure generalization / toy examples
\begin{figure}
     \centering
     \begin{subfigure}[b]{0.22\textwidth}
         \centering
         \includegraphics[width=\textwidth]{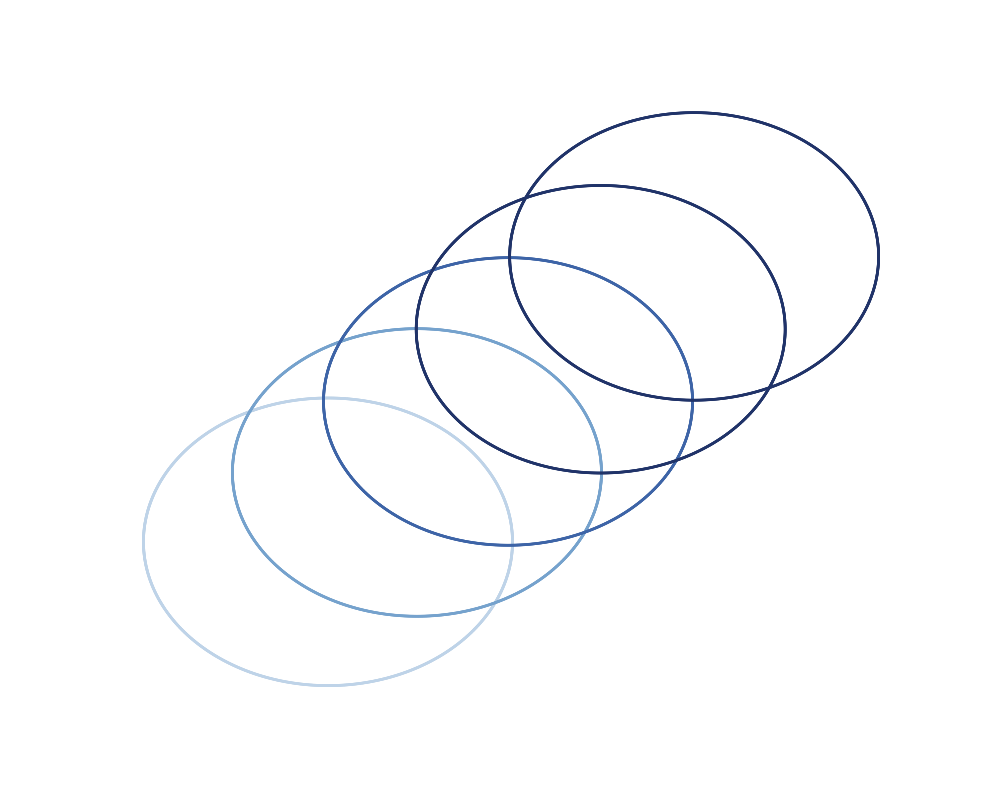}
         \caption{Training set}
         \label{fig:train-circles}
     \end{subfigure}
     \hfill
     \begin{subfigure}[b]{0.22\textwidth}
         \centering
         \includegraphics[width=\textwidth]{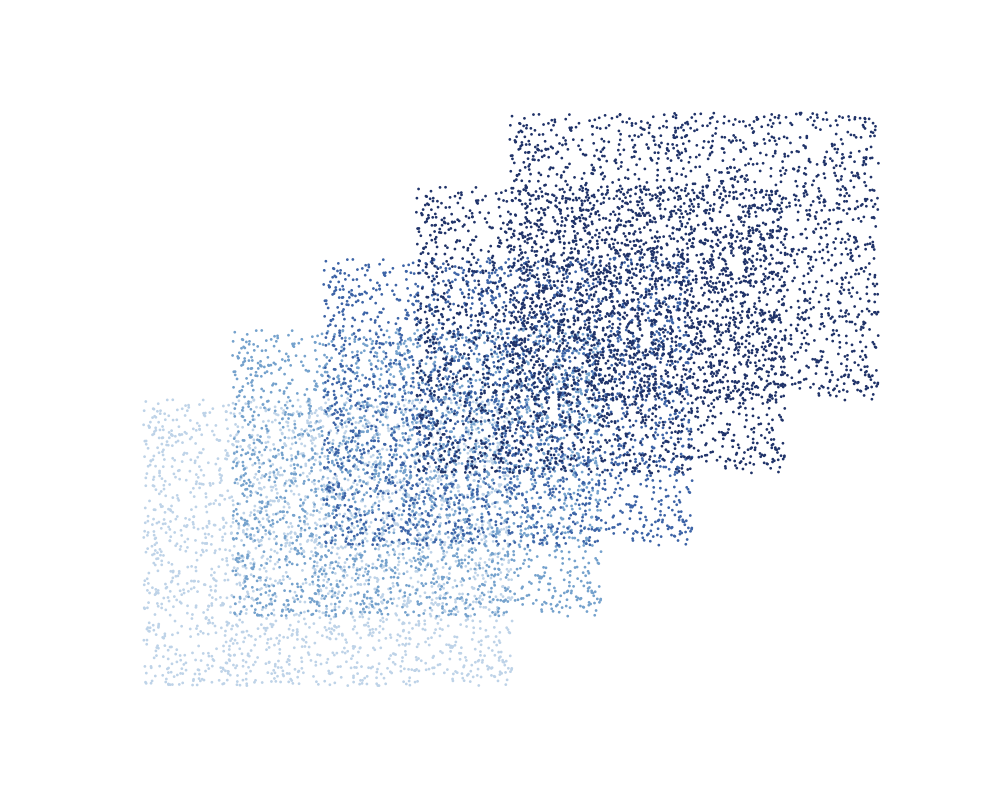}
         \caption{Squares}
         \label{fig:gen-squares1}
     \end{subfigure}
     \hfill
     \begin{subfigure}[b]{0.22\textwidth}
         \centering
         \includegraphics[width=\textwidth]{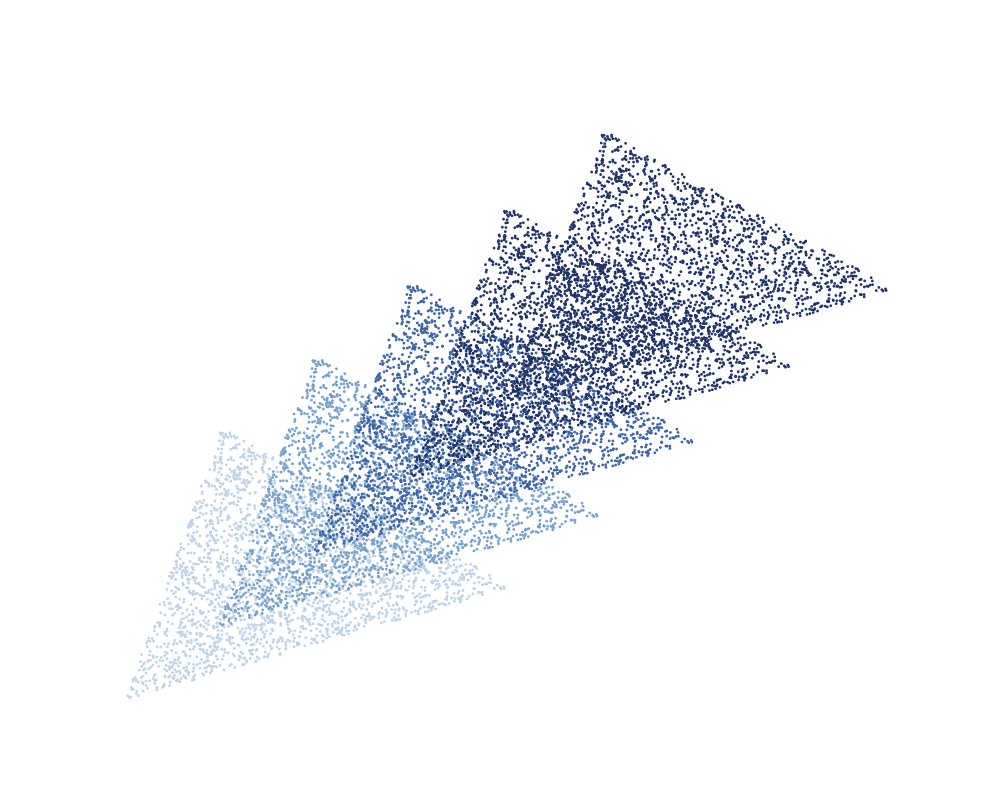}
         \caption{Triangles}
         \label{fig:gen-triangles}
     \end{subfigure}
     \hfill
     \begin{subfigure}[b]{0.22\textwidth}
         \centering
         \includegraphics[width=\textwidth]{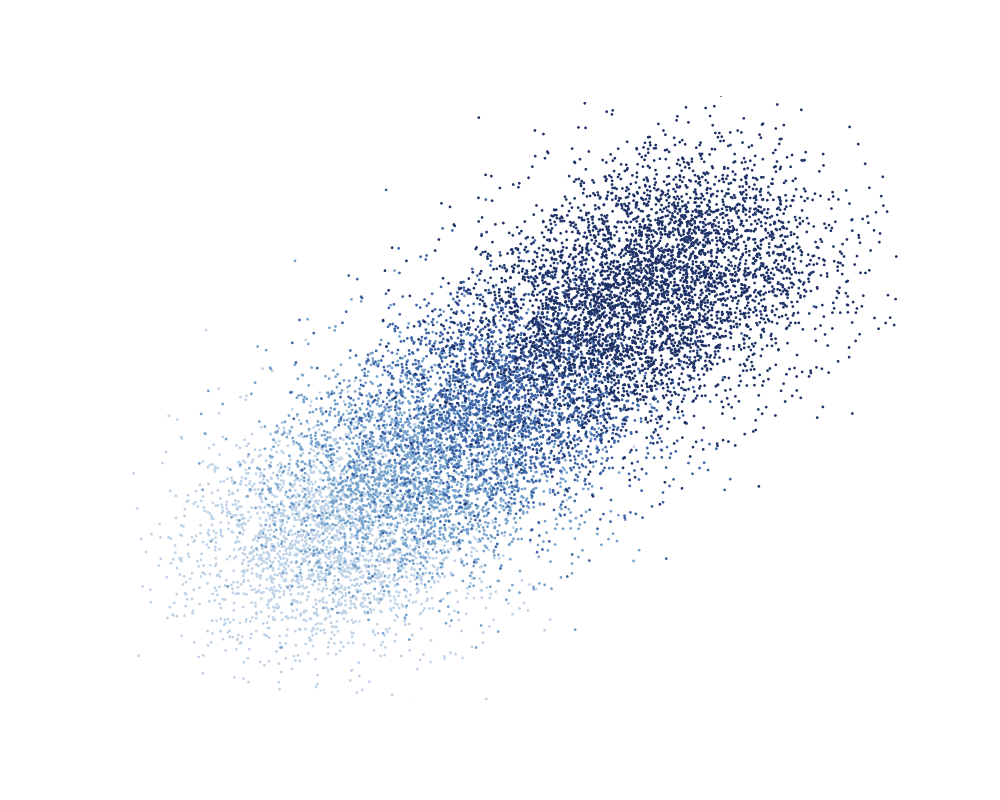}
         \caption{Gaussians}
         \label{fig:gen-gaussians1}
     \end{subfigure}
     \begin{subfigure}[b]{0.22\textwidth}
         \centering
         \includegraphics[width=\textwidth]{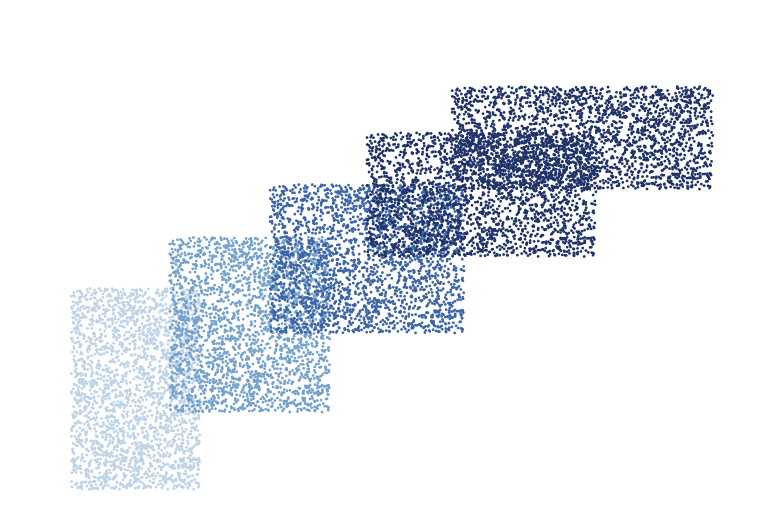}
         \caption{Training set}
         \label{fig:train-rect}
     \end{subfigure}
     \hfill
     \begin{subfigure}[b]{0.22\textwidth}
         \centering
         \includegraphics[width=\textwidth]{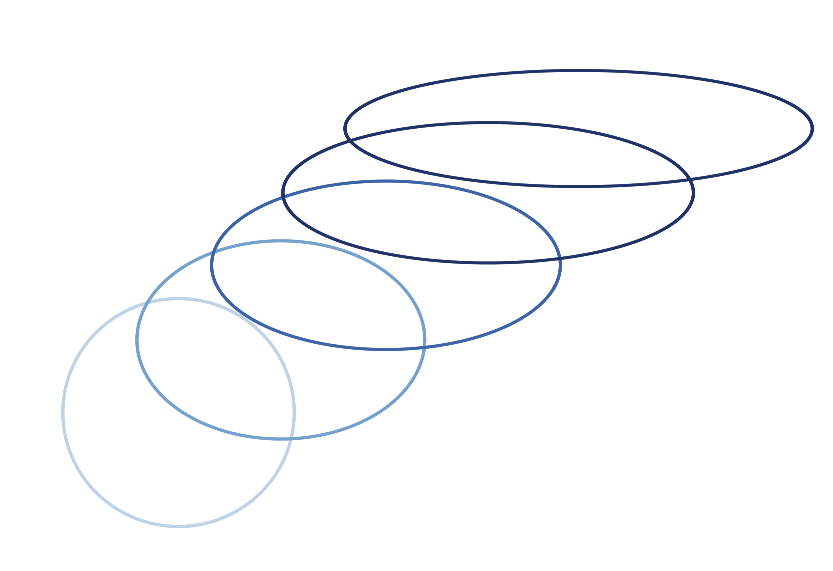}
         \caption{Circles}
         \label{fig:gen-circles}
     \end{subfigure}
     \hfill
     \begin{subfigure}[b]{0.22\textwidth}
         \centering
         \includegraphics[width=\textwidth]{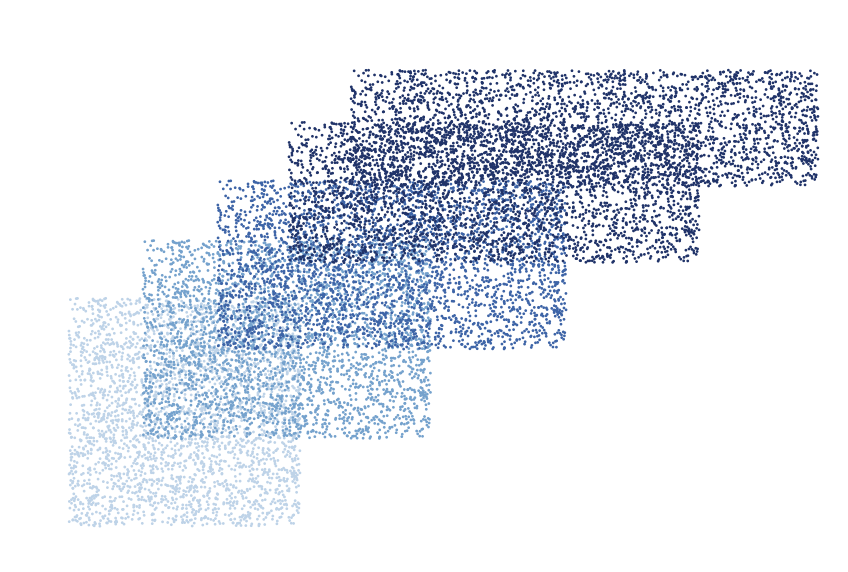}
         \caption{Squares}
         \label{fig:gen-squares2}
     \end{subfigure}
     \hfill
     \begin{subfigure}[b]{0.22\textwidth}
         \centering
         \includegraphics[width=\textwidth]{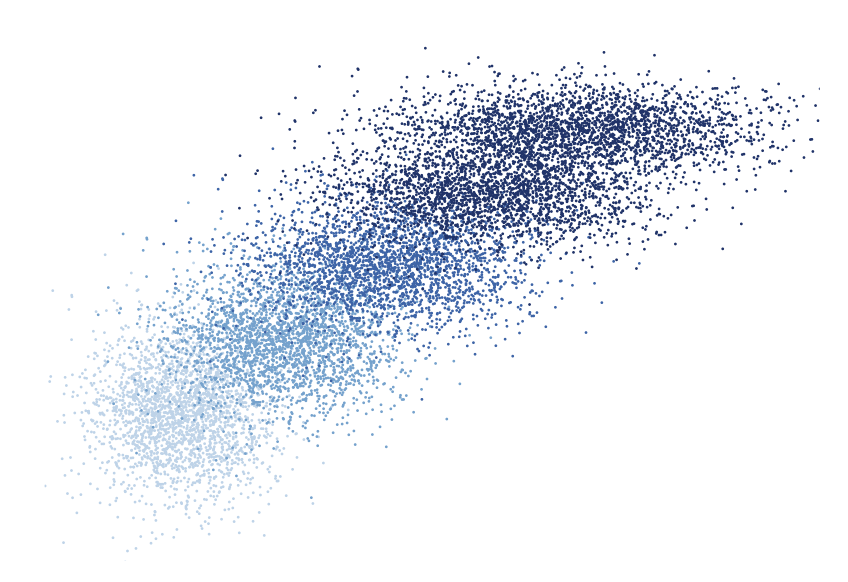}
         \caption{Gaussians}
         \label{fig:gen-gaussians2}
     \end{subfigure}
        \caption{Examples of transported data sets for protocols $\mathcal{P}_1$ (top) and $\mathcal{P}_1$ (bottom).}
        \label{fig:transport-gen}
\end{figure}

Fig. \ref{fig:transport-gen} aims at illustrating the capacity of generalization of the transport map learnt by SWOT-Flow. In this experiment, SWOT-Flow has been trained following the simulation protocols $\mathcal{P}_1$ (Fig. \ref{fig:train-circles}) or $\mathcal{P}_2$ (Fig. \ref{fig:train-rect}). Once trained on the data set associated with each protocol, the NFs are fed with differently shaped data and the elementary transports are monitored as above. Fig \ref{fig:gen-squares1}-\ref{fig:gen-gaussians1} and \ref{fig:gen-squares2}-\ref{fig:gen-gaussians2} show the results when using square-, triangle-, Gaussian-shaped data sets for both protocols, respectively. As expected, all initial distributions are either simply translated in case $\mathcal{P}_1$ or translated and stretched in case $\mathcal{P}_2$. The intermediate distributions correspond to the expected barycenters as well. Fig. \ref{fig:transport-gen} clearly demonstrates the generalization capacity of the proposed approach. % Both approach have been trained with the same number of iterations (3000).

\subsubsection{Multivariate Gaussians with varying dimensions.} When the source and target distributions $\mu$ and $\nu$ of a transportation problem are multivariate Gaussians, the Wasserstein barycenters defined by \eqref{eq:barycenter} are also multivariate Gaussian distributions. In this case, an efficient fixed-point algorithm can be used to estimate its mean vector $\mathbf{a}$ and covariance matrix $\boldsymbol{\Sigma}$ \cite{fixedpoint}. This experiment capitalizes on this finding to assess the ability of SWOT-Flow to approximate Wasserstein barycenters, as stated in Section \ref{subsec:barycenter}. To this end, the algorithm designed in  \cite{fixedpoint} is implemented to estimate the actual barycenter associated with two prescribed  multivariate Gaussian distributions for $\alpha = 1-\alpha = \frac{1}{2}$. This barycenter is compared to the image measure $T_{[m]\sharp} \mu$ estimated by SWOT-Flow with $m = \frac{M}{2}$. More precisely, the mean vector and the covariance matrix of the barycenter are compared to their maximum likelihood estimates $\hat{\mathbf{a}}$ and $\hat{\boldsymbol{\Sigma}}$ computed from the samples $\{T_{[m]}(x_n)\}_{n=1}^N$ transported by the first $m$ flows. The resulting mean square errors (MSEs) 
\begin{eqnarray}
    \mathrm{MSE}(\mathbf{a}) = \left \| {\mathbf{a}} - \hat{\mathbf{a}} \right \|_2^2 \quad \text{and} \quad \mathrm{MSE}(\boldsymbol{\Sigma}) = \| {\boldsymbol{\Sigma}} - \hat{\boldsymbol{\Sigma}} \|_\mathrm{F}^2
\end{eqnarray}
are reported in Table \ref{tab:stato-test} for varying dimensions ranging from $2$ to $8$. This table also reports the MSEs reached by other state-of-the-art free-support methods \cite{fastBarycenters,StochasticBarycenters,continuousBarycenter}. For the methods \cite{fastBarycenters} and \cite{StochasticBarycenters}, $n=5000$ and $n=100$ support points have been used, respectively, since these are the maximum numbers allowed for the algorithms to terminate in reasonable computational times. SWOT-Flow compares favorably to state-of-the-art methods since reported MSEs in Table \ref{tab:stato-test} appear to be most often the smallest. These observation may call for a more general study, but remains noticeable since SWOT-Flow has not been specifically designed to compute the Wasserstein barycenters, contrary to alternate methods.

%The comparison of these performance measures should be conducted with care since SWOT-Flow has not been specifically designed to compute the Wasserstein barycenters, contrary to these alternate methods. 

\begingroup
\setlength{\tabcolsep}{7pt} % Default value: 6pt
\renewcommand{\arraystretch}{1.15} % Default value: 1
\begin{table}
    \centering
    \caption{Performance of the estimation of the median barycenters. Reported scores result from the average  over 5 Monte Carlo runs.}
    \begin{tabular}{|cc|c|cccc|}
    \cline{4-7}
    \multicolumn{3}{c|}{} & \cite{fastBarycenters} & \cite{StochasticBarycenters} & \cite{continuousBarycenter} & SWOT-Flow \\
    \hline
    \multirow{8}{*}{\rotatebox{90}{Dimension}} & \multirow{2}{*}{2} & $\mathrm{MSE}(\mathbf{a}) $  & $9.99\cdot10^{-5}$  & $3.14\cdot10^{-4}$  & $1.17\cdot 10^{-4}$  & $\mathbf{8.09\cdot10^{-5}}$ \\
    & & $\mathrm{MSE}(\boldsymbol{\Sigma}) $ & $7.28\cdot10^{-4}$  & $2.39\cdot10^{-3}$  & $1.98\cdot10^{-3}$  & $\mathbf{1.44\cdot 10^{-4}}$ \\
    \cline{2-7}
    & \multirow{2}{*}{4} & $\mathrm{MSE}(\mathbf{a}) $  & $1.73\cdot10^{-3}$  & $1.68\cdot10^{-3}$  & $1.44\cdot10^{-3}$  & $\mathbf{1.44\cdot 10^{-4}}$ \\
    & & $\mathrm{MSE}(\boldsymbol{\Sigma}) $ & $1.35\cdot10^{-2}$  & $2.50\cdot10^{-2}$  & $1.22\cdot10^{-2}$  &  $\mathbf{3.61\cdot 10^{-4}}$\\
   \cline{2-7}
    & \multirow{2}{*}{6} & $\mathrm{MSE}(\mathbf{a}) $  & $2.04\cdot10^{-3}$  & $\mathbf{2.58\cdot10^{-3}}$  & $3.24\cdot10^{-3}$  & $1.23\cdot10^{-2}$ \\
    & & $\mathrm{MSE}(\boldsymbol{\Sigma}) $ & $4.38\cdot10^{-2}$  & $8.86\cdot10^{-2}$  & $2.37\cdot10^{-2}$  & $\mathbf{5.29\cdot 10^{-4}}$ \\
    \cline{2-7}
    & \multirow{2}{*}{8} & $\mathrm{MSE}(\mathbf{a}) $  & $\mathbf{1.23\cdot10^{-3}}$  & $1.48\cdot10^{-3}$  & $3.14\cdot10^{-3}$  & $1.29\cdot10^{-2}$ \\
    & & $\mathrm{MSE}(\boldsymbol{\Sigma}) $ & $8.31\cdot10^{-2}$  & $1.64\cdot10^{-1}$  & $4.23\cdot10^{-2}$  & $\mathbf{2.22\cdot10^{-3}}$ \\
    \hline
    \end{tabular}
    \label{tab:stato-test}
\end{table}
\endgroup

\subsection{Unsupervised word translation}

In a second set of experiments, the performance of SWOT-Flow has been assessed on the task of unsupervised word translation. Given word embeddings trained on two monolingual corpora, the goal is to infer a bilingual dictionary
by aligning the corresponding word vectors.

\subsubsection{Experiment description.} This experiment considers the task of aligning two sets of points in high dimension. More precisely, it aims at inferring a bilingual lexicon, without supervision, by aligning word embeddings trained on monolingual data. FastText \cite{fasttext} has been implemented to learn the word vectors used for representation. It provides monolingual embeddings of dimension 300 trained on Wikipedia corpora. Words are lower-cased, and those that appear less than 5 times are discarded for training. As a post-processing step, only the first $50$k most frequent words are selected in the reported experiments. 

\subsubsection{Architecture.} The proposed SWOT-Flow method has been implemented using a RealNVP architecture. The scale function $\mathsf{E}_m(\cdot)$ and the offset function $\mathsf{D}_m(\cdot)$ are multilayer neural networks with two hidden layers of size 512 and hyperbolic tangent activation function. Adam has been used as an optimizer with a learning rate of $1\cdot 10^{-3}$. The number of slices involved in the Monte Carlo approximation of the SW distance in \eqref{eq:SW_MonteCarlo} has been progressively increased from $J=500$ slices to $J=3000$ by steps of 50. For each number of slices, 100 epochs have been performed. The hyperparameters $\lambda$ and $\gamma$ adjusting the weights of the composite regularization have been increased by a factor of $5\%$ every steps of 500 slices.

\begingroup
\setlength{\tabcolsep}{3.4pt} % Default value: 6pt
\renewcommand{\arraystretch}{1.15} % Default value: 1
\begin{table}
\centering
\caption{Comparison of accuracies obtained by SWOT-Flow and adv-net \cite{Adversarial} for unsupervised word translation ('en' is English, 'fr' is French, 'de' is German, 'ru' is Russian).}
\begin{tabular}{|cc|cc|cc|cc|cc|}
    \hline
    \multicolumn{2}{|c|}{Method} & \multicolumn{1}{c}{en-es} & \multicolumn{1}{c|}{es-en} & \multicolumn{1}{c}{en-fr} & \multicolumn{1}{c|}{fr-en} & \multicolumn{1}{c}{en-de} & \multicolumn{1}{c|}{de-en} & \multicolumn{1}{c}{en-ru} & \multicolumn{1}{c|}{ru-en} \\
    \hline
    \multirow{2}{*}{{SWOT-Flow}} %& $\widehat{SW}$ ($\cdot 10^{-6}$)& $2.27$ & $3.6$ & $0.95$ & $3.31$ & $1.02$ & $1.35$ & $2.17$ & $2.35$ \\
    & 20-NN & $37.4$ & $24.2$ & $46.6$ & $34.1$ & $44.4$ & $27.6$ & $14.4$ & $3.8$ \\
    & 10-NN & $\mathbf{33.5}$ & $\mathbf{22.5}$ & $\mathbf{42.5}$ & $32.5$ & $39.5$ & $26.8$ & $\mathbf{10.2}$ & $2.1$ \\
    \hline
    \multirow{1}{*}{adv-net \cite{Adversarial}} & 10-NN & $31.4$ & $21.2$ & $39.6$ & $\mathbf{35.1}$ & $\mathbf{40.1}$ & $\mathbf{27.1}$ & $7.1$ & $\mathbf{2.3}$ \\
    \hline
\end{tabular}
\label{tab:translation-bench}
\end{table}
\endgroup

\subsubsection{Main results.} To quantitatively measure the quality of SWOT-Flow, the problem of bilingual lexicon induction is addressed, with the same setting as in \cite{Adversarial}. The same evaluation data sets and codes, as well as the same word vectors have been used. Given an input word embedding ($n=1,\ldots,N$ with $N_{\textrm{test}}=1000$) in a given language, the objective is to assess if its counterpart $T(x_n)$ transported by SWOT-Flow belongs to the close neighborhood of the output word embedding $y_n$ in the target language. The neighborhood $\mathcal{V}(y_n)$ is defined as the set of $K$-nearest neighbors  computed in a cosine similarity sense with $K=10$ or $20$ in dimension 300. The overall accuracy is computed as the percentage of correctly transported input samples. Denoting by $\boldsymbol{1}_{\mathsf{A}}$ the indicator function, i.e., $\boldsymbol{1}_{\mathsf{A}}=1$ if the assertion $\mathsf{A}$ is true and $\boldsymbol{1}_{\mathsf{A}}=0$ otherwise,
\begin{equation}
    \mathrm{accuracy} =  \frac{1}{N_{\textrm{test}}} \sum_{n=1}^{N_{\textrm{test}}} \boldsymbol{1}_{\left\{T(x_n) \in \mathcal{V}(y_n)\right\}}  \times 100 \ \text{(\%)}
\end{equation}

Table \ref{tab:translation-bench} reports the accuracy scores for several pairs of languages. Although SWOT-Flow has not been specifically designed to perform word translation, these results show that its overall performance is on par with the adversarial network (adv-net)  proposed  specifically for this task in \cite{Adversarial}. In particular, SWOT-Flow seems to perform well for translation between languages with close origins.

%%%%%
\renewcommand{\subfigsize}{0.40\textwidth}
\begin{figure}
     \centering
\begin{comment}
        \begin{subfigure}[b]{\subfigsize}
         \centering
         \includegraphics[width=\textwidth]{figures_100dpi/english-spanish.png}
         \caption{english $\rightarrow$ spanish}
         \label{fig:gen-disks}
     \end{subfigure}
     \hfill  
     \begin{subfigure}[b]{\subfigsize}
         \centering
         \includegraphics[width=\textwidth]{figures_100dpi/english-russian.png}
         \caption{english $\rightarrow$ russian}
         \label{fig:english-russian}
     \end{subfigure}
\end{comment}    
     \begin{subfigure}[b]{\subfigsize}
         \centering
         \includegraphics[width=\textwidth,height=0.80\textwidth]{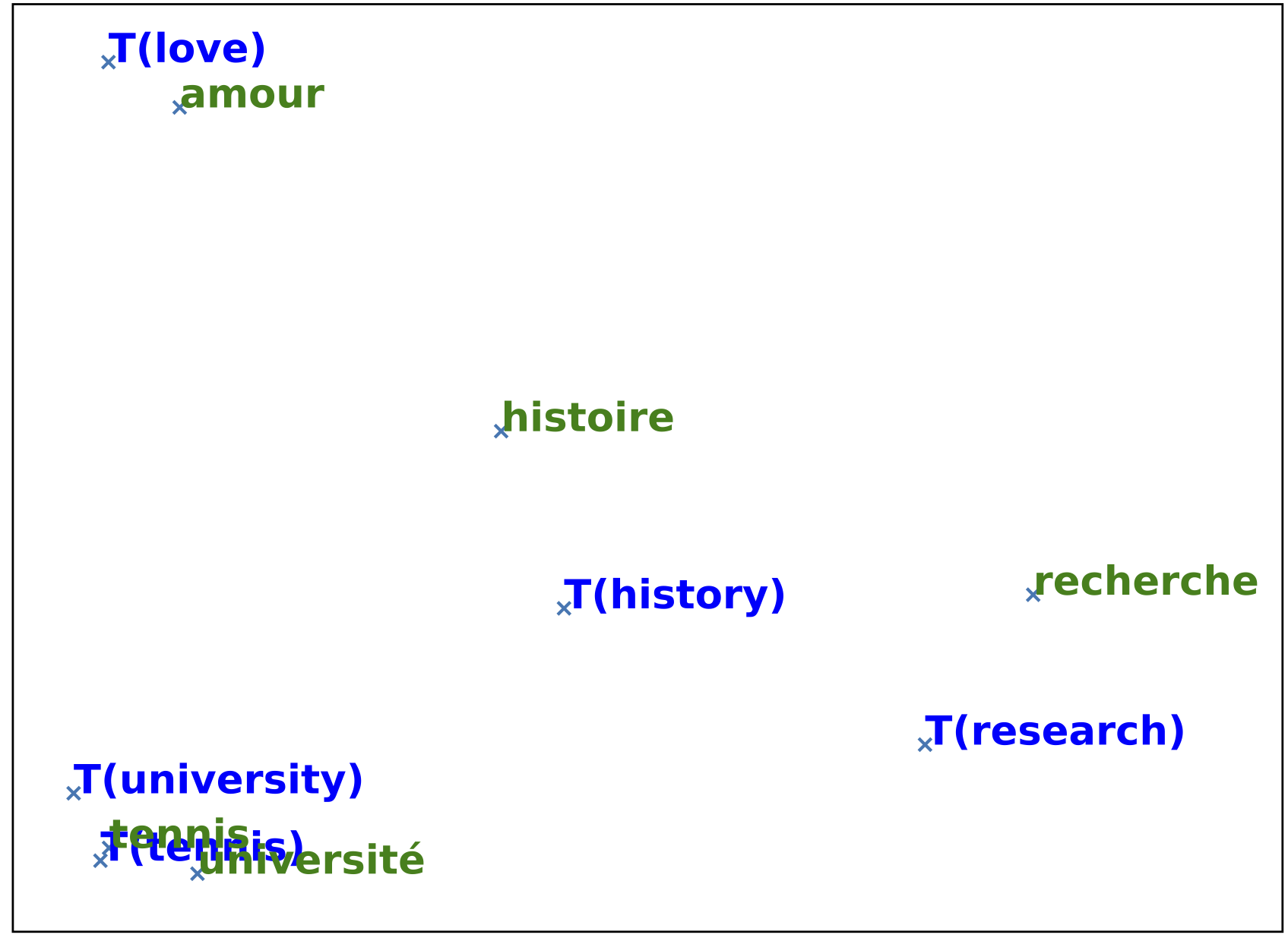}
         \caption{english $\rightarrow$ french}
         \label{fig:english-french}
     \end{subfigure}%
     %\hfill
     \begin{subfigure}[b]{\subfigsize}
         \centering
         \includegraphics[width=\textwidth,height=0.80\textwidth]{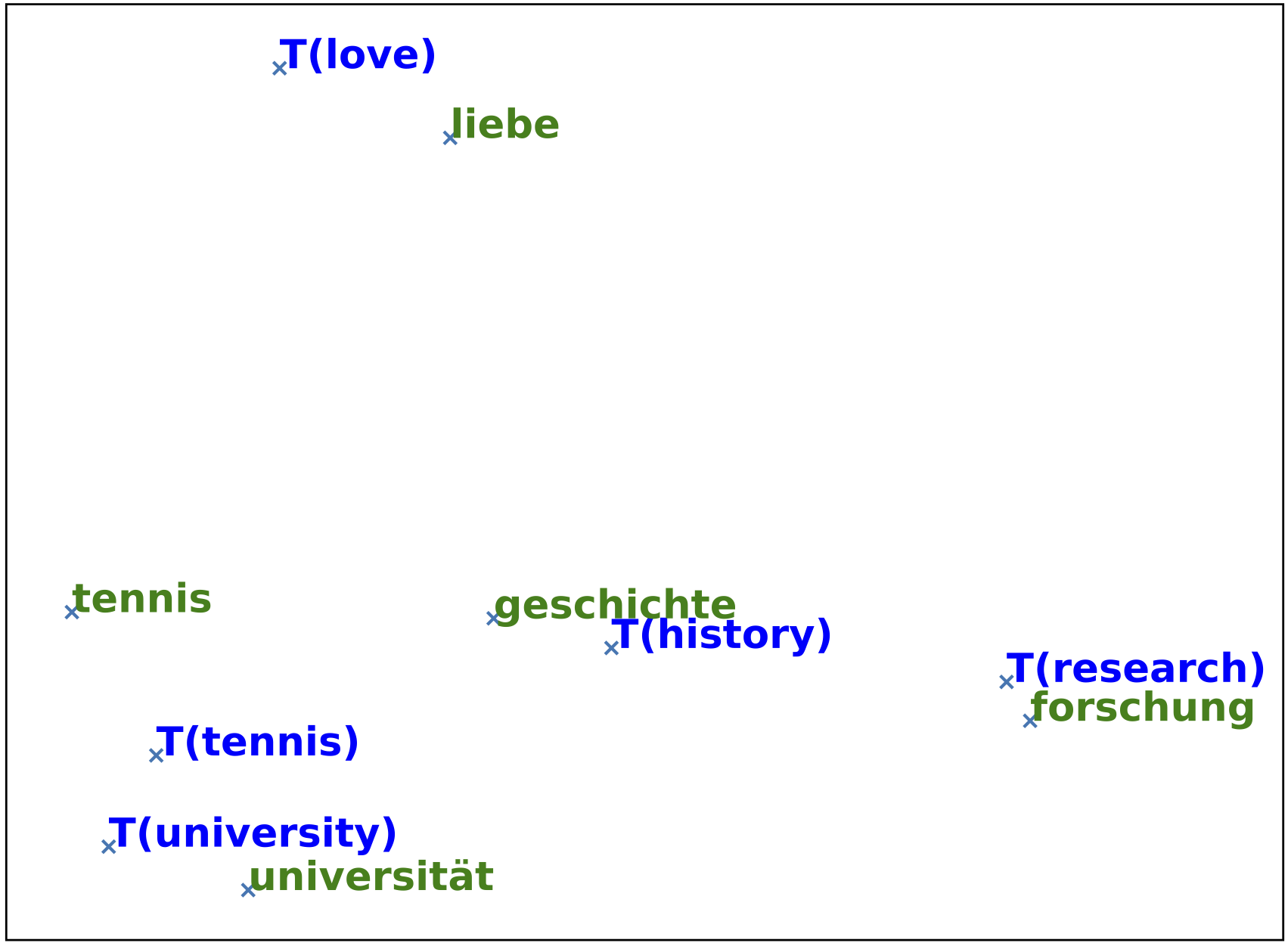}
         \caption{english $\rightarrow$ german}
         \label{fig:gen-squares}
     \end{subfigure}%
         \caption{2D PCA representation of the target word embedding space: the targeted translated (in green) and the transported source (in blue) embedded words.}
        \label{fig:word-embbeding}
\end{figure}

Fig.~\ref{fig:word-embbeding} qualitatively illustrates this good performance by showing how close a set of translated words $T(x_n)$ are to their true translation $y_n$. This representation is obtained by a classical projection on the 2 first PCA components of the target embedded space. The translation of 5 specific words from English to  French or German fall in the close vicinity of their true counterparts.

\section{Discussion}

%Since using Sliced-Wasserstein distance permits to learn a bijective mapping $T$ between any complex distributions, classical normalizing flows model can be extended to other applications such as image-to-image translation, unsupervised domain adaptation \cite{domain-adaptation-OT}. They can be used to simultaneously interpolate across the various domains using the learnt transport map $T$ and $T^{-1}$.

\subsubsection{Cycle consistency.} Cycle consistency, as proposed in CycleGAN \cite{CycleGAN}, aims at learning  meaningful cross-domain mappings such that the data translated from the domain $\mathcal{X}$ to the domain $\mathcal{Y}$ via $T_{\mathcal{X} \rightarrow \mathcal{Y}}$ can be mapped back to the original data points in $\mathcal{X}$ via $T_{\mathcal{Y} \rightarrow \mathcal{X}}$. That is, $T_{\mathcal{Y} \rightarrow \mathcal{X}}\circ T_{\mathcal{X} \rightarrow \mathcal{Y}}(x) \approx x$ for all $x \in \mathcal{X}$. For CycleGan, and many other domain transfer models such as \cite{CycleGAN-OT}, this key property should be enforced by including a cycle consistency term into the loss function. Conversely, since NF-based generative models learn bijective mappings, the proposed SWOT-Flow inherits the cycle consistency property by construction.

\subsubsection{Semi-discrete formulation.} The proposed SWOT-Flow framework has been explicitly derived to approximate OT between two discrete empirical distributions. It can be instanciated to perform semi-discrete OT, i.e., to handle the case where one of distribution is not described by data points but rather given as an explicit continuous probability measure. Instead of relaxing the Monge formulation \eqref{monge} as in \eqref{relaxed}, it would consist in replacing the SW distance with a log-likelihood term $\log f_\nu(\cdot)$ associated with the target continuous measure. The loss function  in \eqref{eq:loss_function} would be replaced by 
\begin{equation}
- \sum_{n=1}^N \log f_\nu(T_{\boldsymbol{\Theta}}({x_n})) + \sum_{n=1}^{N} \sum_{m=1}^{M}  \bigg[ \lambda c(T_{m-1}(x_n), T_m(x_n)) + \gamma \left|{J_{T_m}}(x_n)\right|^{2} \bigg]
%\lambda \int_{\mathcal{X}} c(x, T(x)) \mathrm{d} \mu(x)\right\}
\end{equation}
where the log-likelihood term is evaluated at the data points $\left\{T_{\boldsymbol{\Theta}}({x_n})\right\}_{n=1}^N$ transported by the NF.

\subsubsection{NF to approximate barycenters.} As discussed in Section \ref{subsec:barycenter} and experimentally illustrated in Section \ref{sec:toy_examples}, the flow-based architecture of the SWOT-Flow network leads to intermediate transports, that can be related to Wasserstein barycenters. On the toy Gaussian example considered in Section \ref{sec:toy_examples}, SWOT-Flow provides good approximation of the barycenters, although it has not been specifically designed to perform this task. If one is interested in devising a NF approximating these barycenters, the definition \eqref{eq:barycenter} would lead to the optimization problem
% the proposed approach could be explicitly trained to target Wasserstein barycenters as well. Indeed, we could reconsider the problem \eqref{relaxed} and train intermediate transport $T_[m](\cdot)$ to target a barycenter of weight $\alpha_m = \frac{m}{M}$. The problem would write
\begin{equation}
    \inf_{T}  \left\{\sum_{m=1}^M \alpha_m W_p(\mu,T_{[m]\sharp}\mu) + (1-\alpha_m) W_p(T_{[m]\sharp}\mu,\nu)\right\}
    \label{eq:barycenter-loss}
\end{equation}
with $\alpha_m = \frac{m}{M}$. When handling empirical measures described by samples, the subsequent discretization would require to replace both terms with Monte Carlo approximations \eqref{eq:SW_MonteCarlo} of the SW distances. However, this would lead to a computationally demanding training procedure.

%%  à discuter (si on a le temps...)

\section{Conclusion}\label{sec:conclusion}

We propose a new method to learn the optimal transport map between two empirical distributions from sets of available samples. To this aim, we write a relaxed and penalized formulation of the Monge problem. 
This formulation is used to build a loss function that balances between the cost of the transport and the proximity in Wasserstein distance between the transported base distribution and the target one. 
The proposed approach relies on normalizing flows, a family of invertible neural networks. Up to our knowledge, this is the first method that is able to learn such a generalizable transport operator. 
As a side benefit, the multiple flow architecture of the proposed network interestingly yields intermediate transports and Wasserstein barycenters.
The proposed method is illustrated by numerical experiments on toy examples as well as an unsupervised word translation task. 
Future work will aim at extending these results to high dimensional applications.

%The main contribution of this paper is a use of normalization flows to learn an approximation of an optimal transport plan between two empirical distributions. We propose a relaxed and penalized formulation of the Monge problem. This formulation is used as a cost function to train a bijective network whose architecture allows access to intermediate transports, which can be associated with Wasserstein barycentres. 

\begin{comment}
\section*{Acknowledgements}
This work was partly supported by the ANR project « Chaire IA Sherlock »  ANR-20-CHIA-0031-01 hold by P. Chainais, as well as by the national support within the PIA ANR-16-IDEX-0004 ULNE and Region HDF.
\end{comment}

\bibliographystyle{splncs04}
\bibliography{strings_all_ref,reference}

%\printbibliography

\end{document}